\pdfoutput=1

\documentclass[11pt]{article}

\usepackage{acl}

\usepackage{times}
\usepackage{latexsym}

\usepackage[T1]{fontenc}

\usepackage[utf8]{inputenc}

\usepackage{microtype}

\usepackage{tabularx}
\usepackage{bm}
\usepackage{amsmath}
\usepackage{mathrsfs}
\usepackage{multirow}
\usepackage{booktabs}
\usepackage{tikz}
\usepackage{tikz-qtree}
\usepackage{color}
\usepackage{xcolor}
\usepackage{makecell}
\usepackage[switch]{lineno}
\usepackage{enumitem}
\usepackage{CJKutf8}
\usepackage{makecell}

\title{Faster and Better Grammar-based Text-to-SQL Parsing 
via Clause-level Parallel Decoding and Alignment Loss}

\author{
Kun Wu$^1$,
Lijie Wang$^1$,
Zhenghua Li$^2$,
Xinyan Xiao$^1$ \\
1. Baidu Inc, Beijing, China \\
2. Institute of Artificial Intelligence, School of Computer Science and Technology,\\
Soochow University, Suzhou, China \\
\{wukun04,wanglijie,xiaoxinyan\}@baidu.com \\
zhli13@suda.edu.cn\
}

\begin{document}
\begin{CJK}{UTF8}{gkai}

\maketitle

\begin{abstract}

Grammar-based parsers have achieved high performance in the cross-domain text-to-SQL parsing task, but suffer from low decoding efficiency due to the much larger number of actions for grammar selection than that of tokens in SQL queries. 
Meanwhile, how to better align SQL clauses and question segments has been a key challenge for parsing performance. 
Therefore, this paper proposes clause-level parallel decoding and alignment loss to enhance two high-performance grammar-based parsers, i.e., RATSQL and LGESQL.
Experimental results of two parsers show that our method obtains consistent improvements both in accuracy and decoding speed. 
\end{abstract}
\section{Introduction}

\begin{figure*}[tb]
\centering
\includegraphics[width=0.95\textwidth]{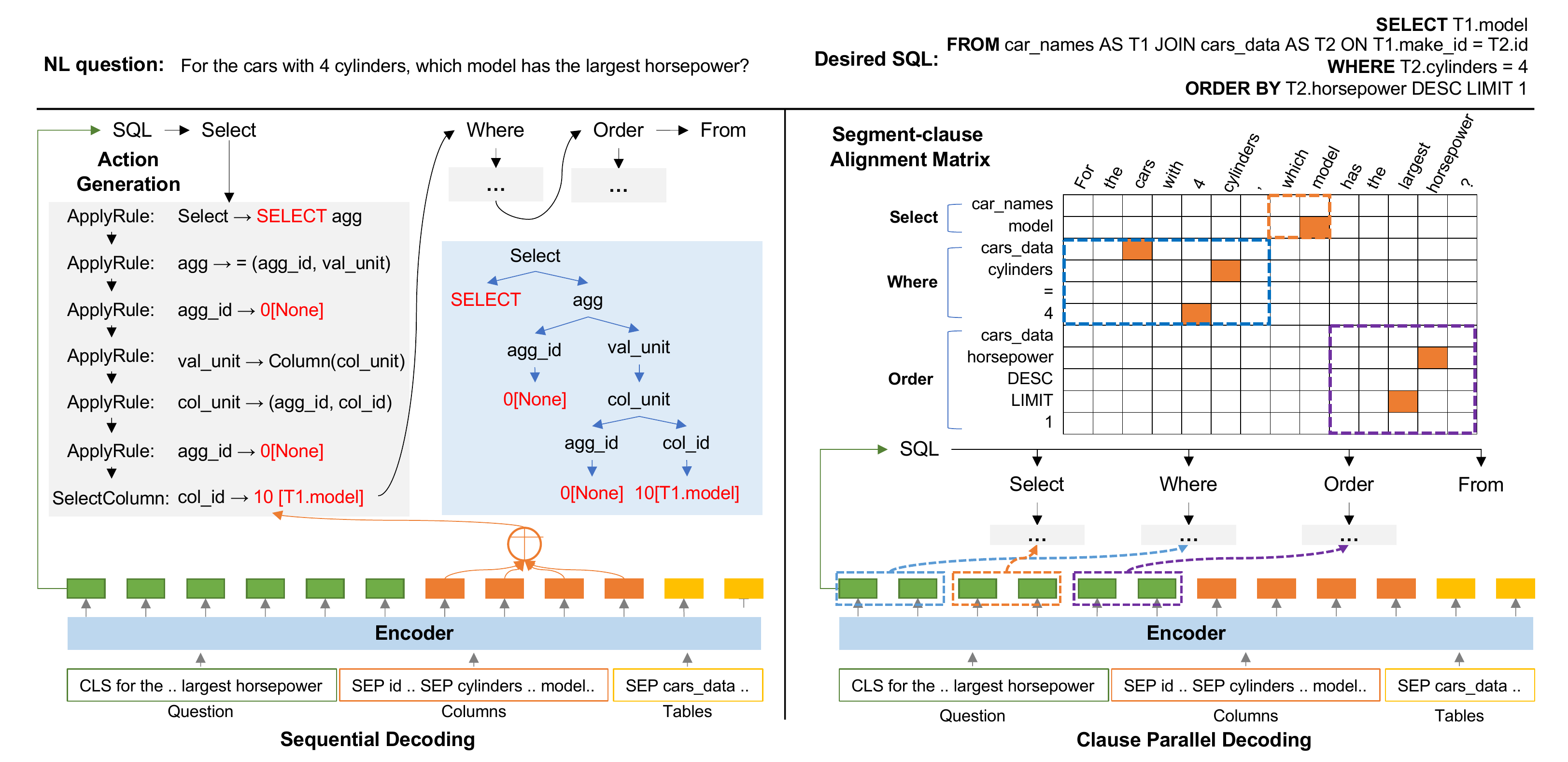}
\caption{An overview of our approach. The left side shows the generation process of sequential decoding in RATSQL grammar-based decoder, and the right side gives our proposed parallel decoding, where all clauses are generated independently. Meanwhile, according to alignments between SQL clauses and question segments, as shown by the segment-clause alignment matrix, a clause-level alignment loss is incorporated during training.}
\label{fig:framework}
\end{figure*}

Text-to-SQL parsing aims to automatically transform natural language (NL) questions into SQL queries based on the given databases (DBs) \cite{tang2001using}, as depicted at the top of Figure \ref{fig:framework}.  
Recently, several high-quality cross-domain text-to-SQL datasets have been released, strongly boosting the research interest and progress in this task \cite{zhong2017seq2sql, yu2018spider, wang-etal-2020-dusql}. 
Most early works generate SQL queries in a token-level seq2seq manner \cite{zhong2017seq2sql,dong2018coarse}, or by filling DB elements into SQL slots \cite{xu2017sqlnet,yu2018typesql}, both of which are known as token-based parsers.
In contrast, a grammar-based parser incorporates SQL grammar into the decoder to guarantee the grammaticality of output SQL queries \cite{yin2018tranx}, including RATSQL \cite{wang2020rat} and LGESQL \cite{cao2021lgesql}, both of which have achieved state-of-the-art (SOTA) performance on complex datasets. They share the same decoder with different grammars, and LGESQL further uses a new graph encoder to enhance presentations of the question words and 
DB schema items.

Concretely, a grammar-based parser builds a tree complying with SQL grammar via a sequence of actions, as shown on the left of Figure \ref{fig:framework}, where the tree's leaf nodes form the final SQL query. 
In spite of its high performance, the number of actions is usually much larger than the number of tokens in the SQL query, due to the generation of non-leaf nodes. 
This makes the decoding process extremely inefficient. 
To alleviate the inefficiency issue, DuoRAT \cite{scholak2021duorat} uses a transformer-based decoder to replace the parent-feeding LSTM decoder in RATSQL \cite{wang2020rat}, which can improve the training efficiency given gold-standard SQL queries. 
Unfortunately, their method does not influence the testing speed, which is very important in real applications.

As discussed by many previous works, one characteristic of the text-to-SQL task is that an SQL clause usually depends on a local segment of the input question \cite{zeng2020recparser,yin2021compositional,wu2021data}.   
Recent works try to exploit alignments between SQL clauses and question segments for better handling some specific SQL structures. 
\citet{zeng2020recparser} propose a recursive parsing framework that can elegantly generate complicated nested SQL queries. The basic idea is explicitly encouraging the decoder to focus on different question segments when generating different nested layers. 
Based on a token-based parser, \citet{yin2021compositional} incorporate an extra attention loss to capture such alignments, 
which is proved to be helpful for dealing with compositional SQL queries. 

To handle the above two issues, we propose to enhance grammar-based parsers via clause-level parallel decoding and alignment loss.
First, we propose to generate SQL clauses in parallel, that is, clauses are generated independently of each other and simultaneously.
Second, we propose a clause-level alignment training loss to encourage the model to focus on only related question segment when generating a clause. 
We implement these two strategies based on two SOTA grammar-based parsers, RATSQL and LGESQL. 
Experimental results on Spider show that our methods obtain consistent improvements both in accuracy and testing speed. We will release our code at \url{https://github}.

\section{Our Proposed Model}
\label{sec:method}

\subsection{Grammar-based Text-to-SQL Parsing}
\label{ssec:grammar_decoder}

As shown on the left side of Figure \ref{fig:framework}, the decoder generates SQL queries via generating actions to select grammar rules in the depth-first search order.
Specifically, there are three types of actions, i.e., \text{ApplyRule}, \text{SelectColumn} and \text{SelectTable}. 
\text{ApplyRule}($r$) applies an action $r$ to expand the focus node, and is used to gradually create a skeleton tree without concrete DB elements. 
\text{SelectColumn}($c$) and \text{SelectTable}($t$) are used to fill a skeleton tree with concrete values by selecting a column name $c$ or a table name $t$, respectively.

We take an example in Figure \ref{fig:framework} to illustrate the process of grammar-based decoder. Suppose that the decoder is at the ``agg'' node (current node, denoted as $n_i$) under the ``Select'' node (father node, denoted as $f_i$), and the next action is ``\text{ApplyRule}(agg $\rightarrow$ agg\_id val\_unit)''. The LSTM state of the decoder is updated as follows.
\begin{equation}\label{equation:lstm}
\begin{split}
& \mathbf{c}_{t+1}, \mathbf{h}_{t+1} = \texttt{LSTM}(\mathbf{c}_{t},\mathbf{h}_{t}, \mathbf{i}_{t}) \\
& \mathbf{i}_t = [\mathbf{e}_{a_t};\mathbf{e}_{n_i};\mathbf{e}_{\texttt{type}(n_i)};\mathbf{z}_t;\mathbf{h}_{\texttt{tm}(f_i)}]
\end{split}
\end{equation}
where $\mathbf{c}_t$ and $\mathbf{h}_t$ are the cell state and the output vector at step $t$; $\mathbf{e}_n$ represents the embedding vector of the input $n$; $a_{t}$ denotes the previous action; $\texttt{type}(.)$ returns the type of a node\footnote{The parser assigns a type for each node according to its role in SQL, such as ``agg'' for aggregations.}; $\mathbf{z}_t$ is the contextual representation vector after attending to the encoder outputs; $\texttt{tm}(f_i)$ denotes the timestamp when $f_i$ has been just generated. 

\subsection{Clause-level Parallel Decoding}
\label{ssec:clause_decoding}

During decoding, the grammar-based parser actually generates a SQL query by sequentially creating clauses (seeing Table \ref{tab:six_clauses}) in a predefined order, as shown on the left side of Figure \ref{fig:framework}.
For instance, after completing the SELECT clause, the parser tries to expand the WHERE clause. If ``WHERE $\rightarrow$ None'' is selected by the decoder, it means that the final SQL query does not include a WHERE clause and the decoder will move on to generate the GROUP clause, and so on. 
In fact, the generation of different clauses is quite loosely connected.
This motivates us that we may generate all SQL clauses independently and in parallel via batch processing, which obviously can improve decoding efficiency. 

Specifically, major differences between parallel and sequential decoding lie in the initial LSTM state of each clause, reflected in $\mathbf{c}_0$, $\mathbf{h}_0$, and the previous action $a_{0}$ in Equation \ref{equation:lstm}.
In sequential decoding, the initial status for a subsequent clause is inherited from and thus depends on the previous clause. 
In contrast, in parallel decoding, each clause has the same initial status, which we believe is more reasonable considering the loose dependency between adjacent clauses. 

\begin{table}[tb]
\renewcommand\tabcolsep{3pt}
\renewcommand{\arraystretch}{1.5}
\centering
\begin{small}
\begin{tabular}{l l}
\hline Clauses & Production Rules \\
\hline
SELECT & SELECT agg $\mid$ SELECT agg, agg $\mid$ ...\\
WHERE & WHERE $\mid$ None \\
GROUP & GROUP\_BY $\mid$ GROUP\_BY \& HAVING $\mid$ None \\
ORDER & ORDER\_BY $\mid$ ORDER\_BY \& LIMIT $\mid$ None \\
IEU & INTERSECT $\mid$ EXCEPT $\mid$ UNION $\mid$ None \\
FROM & FROM Table1 $\mid$ FROM Table1, Table2 $\mid$ ...\\
\hline
\end{tabular}
\end{small}
\caption{Six common types of clauses  used for parallel decoding in our work.}
\label{tab:six_clauses}
\end{table}

\subsection{Clause-level Alignment Loss}
\label{ssec:clause_align}

Figure \ref{fig:framework} shows the alignment between question segments and SQL clauses. We use this alignment to improve clause generation by introducing a clause-level alignment training loss to encourage the model to focus on the aligned question segment in the clause generation.

\textbf{Clause-level Alignment Acquisition.} Given a question/SQL pair, for each DB element and condition value in the SQL query, we search for some tokens from NL question to align them, so as to get a token-level alignment matrix. In this process, we use the string-matching method which is commonly used for token-level schema linking in recent works \cite{guo2019towards, wu2021data}. As shown in the alignment matrix in Figure \ref{fig:framework}, token-level alignments are marked in orange box. 

Then we use a simple heuristic algorithm to extract a question segment for each SQL clause from existing token-level alignment results.
For each clause, we take the shortest question segment that contains all DB elements and values in the clause as its aligned segment. As shown in the alignment matrix of Figure \ref{fig:framework}, the question segment for a clause is marked by a dashed bounding box. Please note that the question segments for different clauses may have overlaps. Finally, there are about 23\% question/SQL pairs missing segment-clause alignments. For these pairs, we align each clause to the whole question. We believe that higher-quality alignment  may lead to higher gains, which we leave as future work.

\textbf{Clause-level Alignment Loss.} After aligning SQL clauses with NL question segments, we design an extra training loss to inject such clause-level alignment into the parsing model. 
Intuitively, the model can be benefited by paying more attention to related aligned segments during clause generation. 

In our grammar-based parser, a clause is generated by a sequence of actions.
For instance, the SELECT clause in Figure \ref{fig:framework}, i.e., ``select T1.model'', which is aligned to ``which model'', is generated by six \text{ApplyRule} actions and one \text{SelectColumn} action. 
For each \text{ApplyRule}($r$) action, we define a prior token-wise alignment probability towards its corresponding segment\footnote{We don't use alignment loss for other two actions, since they tend to be closely related with one or two tokens in NL question. Forcing such action to align with too many tokens in a segment degrades the performance, which has been proved by our early-stage preliminary experiments.}. 
Concretely, each token in the segment obtains an averaged probability, whereas tokens outside the segment receive zero. 
\begin{equation}
\centering
\small
P_\texttt{align}(x_i|r_j) = \left\{
\begin{array}{lcl}
0 &  & x_i \notin S\\
1/\texttt{len}(S) &  & x_i \in S \\ 
\end{array}
\right. \nonumber
\end{equation}
where $r_j$ is the rule in the \text{ApplyRule} action, and $x_i$ is the $i$-th token in the sentence, and $\texttt{len}(S)$ is the number of tokens in the aligned segment $S$. 

Then, we define an attention probability from the current decoder state to each question token as
\begin{equation}
\small
\centering
P_\texttt{att}(x_i|r_j) = \texttt{softmax}(..., \mathbf{h_{t}} W_m \mathbf{m_i}, ...) \nonumber
\end{equation}
where $t$ is the timestamp when executing \text{ApplyRule}($r_j$);
$\mathbf{h_{t}}$ is the time $t$'s hidden state of LSTM decoder; 
$\mathbf{m_i}$ is the output vector for $x_i$ in encoder outputs; $W_m$ is a learned matrix.

Finally, we define the alignment loss as the squared distance between the aligned (prior) and attention (modeling) probabilities. 
\begin{equation}
\small
\centering
L = \sum_{j}\sum_{i}(P_\texttt{align}(x_i|r_j)-P_\texttt{att}(x_i|r_j))^2 \nonumber
\end{equation}

In this way, we hope the model learn to attend to certain related question tokens for the sake of better rule selection.

\section{Experiments}
\label{sec:experiment}

\textbf{Dataset.} We conduct experiments on Spider\footnote{\url{https://yale-lily.github.io/spider}}, a complex and cross-domain text-to-SQL dataset. We follow the original data splitting and use the exact matching (EM) accuracy as the evaluation metric. In our experiments, we use the corrected development set released on June 7, 2020.

\textbf{Implementation.} 
We implement our proposed strategies on RATSQL and LGESQL.
The final loss for the training model is the summation of the original loss and our proposed alignment loss.
We set beam size as 1 to evaluate testing speed, and use default values for other parameters.

\defcitealias{choi2021ryansql}{Choi2021}
\defcitealias{guo2019towards}{Guo2019}
\defcitealias{wang2020rat}{Wang2020}
\defcitealias{yu2020grappa}{Yu2020}
\defcitealias{shi2021learning}{Shi2021}
\defcitealias{scholak2021duorat}{Duo2021}
\defcitealias{cao2021lgesql}{Cao2021}
\defcitealias{clark2020electra}{Ele20}

\begin{table}[tb]
\small
\center
\scalebox{0.90} {
\begin{tabular}{l | l | l}
\hline
Models & \makecell[c]{EM \\ Accuracy} & \makecell[c]{Parsing Speed \\ (query/second)}\\
\hline
DuoRAT + BERT (\citetalias{scholak2021duorat}) & 69.9 & - \\
RATSQL (\citetalias{wang2020rat}) & & \\ 
\quad + BERT  & 69.7 & - \\
\quad + GRAPPA (\citetalias{yu2020grappa}) & 73.4 & - \\

LGESQL (\citetalias{cao2021lgesql}) & & \\
\quad + BERT & 73.5 & - \\
\quad + GRAPPA & 74.1 & - \\
\quad + ELECTRA & 75.1 & - \\
\hline
\multicolumn{3}{c}{\textbf{RATSQL}} \\
Orig. + BERT (rerun)  & 71.1$_{\pm0.4}$ & 7.48 \\
Ours + BERT & 72.5$_{\pm0.1}$ & 9.14 (+18.4\%) \\
\quad  w/o Align & 71.7$_{\pm0.2}$ & 9.21 (+18.9\%) \\
\quad  w/o Parallel & 72.4$_{\pm0.1}$ & - \\
Ours + GRAPPA & 74.2$_{\pm0.4}$ & - \\
\hline
\multicolumn{3}{c}{\textbf{LGESQL}} \\
Orig. + ELECTRA (rerun) & 75.1$_{\pm0.7}$ & 11.69 \\
Ours + ELECTRA & 75.7$_{\pm0.6}$ & 15.81(+35.2\%) \\
\quad  w/o Align & 75.3$_{\pm 0.6}$ & 15.84(+35.5\%) \\
\quad  w/o Parallel & 75.6$_{\pm 0.4}$ & - \\
\hline
\end{tabular}
}
\caption{EM accuracy and testing speed on Spider dev set. For our models, we report mean and variance over three runs.}
\label{tab:main_results}
\end{table}

\begin{table}[tb]
\small
\center
\scalebox{0.95}{
\begin{tabularx}{0.48\textwidth}{l|c|c|c|c}
\hline RATSQL & Easy & Medium & Hard & Extra Hard \\
\hline
Orig. + BERT & 87.9 & 72.9 & 63.2 & 49.4\\
Ours + BERT & 88.7 & 74.9 & 64.9 & 50.0\\
\hline
\end{tabularx}
}
\caption{EM accuracy on different hardness levels.}
\label{tab:hard_spider}
\end{table}

\textbf{Main results.} Table \ref{tab:main_results} shows the main results. 
In the first major row, we select several high-performance grammar-based parsers from the Spider leaderboard. 
We report our results in the second and third major rows\footnote{We have submitted our models for obtaining results on test set. However, due to some environmental problems and limited GPU resources, the results have not been returned.}. 
Besides using BERT \cite{devlin2019bert}, we also give the results with GRAPPA, a task-specified pre-trained model. For LGESQL, we give results with ELECTRA \cite{clark2020electra}, which achieves SOTA performance. In order to avoid the effect of performance vibrations, we run each model for 3 times with different random initialization seeds, then report the averaged EM accuracy and the variance. 

\emph{Parallel decoding}. The parallel decoding achieves an average accuracy improvement of 0.6\% and 0.2\% for RATSQL and LGESQL, which proves that there is no strong generation dependency between SQL clauses. 
Meanwhile, the parallel decoding improves parsing speed by 18.9\% and 35.5\% for two parsers, as shown in the last column of Table \ref{tab:main_results}. For LGESQL, the parallel decoding achieves a larger improvement in parsing speed, as its grammar is simpler and the action sequence for each clause is much shorter. The parallel decoding is more effective when there is little difference in action sequence length for all clauses.

\emph{Effect of alignment loss}. The clause-level alignment loss can improve two models by 1.3\% and 0.5\%, although its incorporation slightly decreases the testing speed.
Integrating the two strategies bring slight improvements compared with alignment loss. We suspect that the clause-level alignment loss has weakened the generation dependency between clauses by encouraging the model to focus on the aligned segments when generating clauses.

\begin{figure}[tb]
\centering
\includegraphics[width=0.48\textwidth]{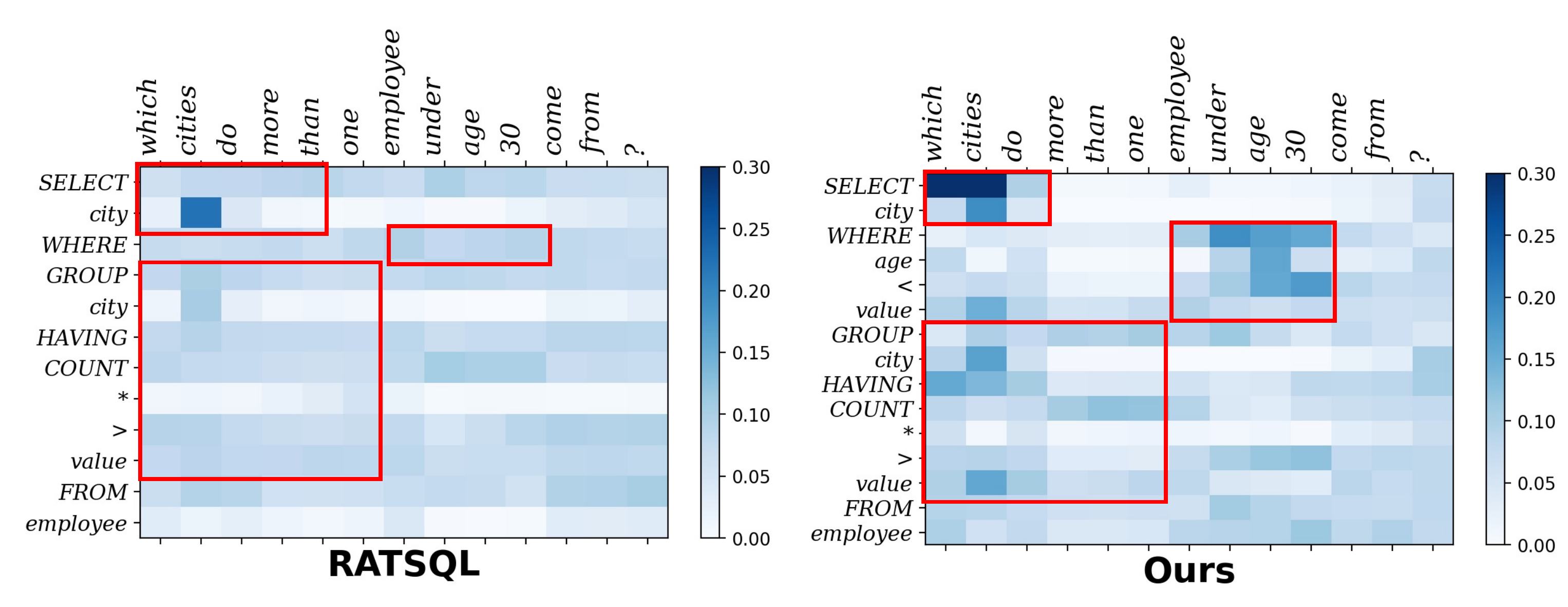}
\caption{Visualization of RATSQL attention scores. Red rectangles highlight alignment blocks that obtain high scores in our model but low scores in baseline.}
\label{fig:matrix}
\end{figure}

\textbf{Hardness analysis}. As shown in Table \ref{tab:hard_spider}, our model outperforms original RATSQL on all hardness levels. 
\citet{yu2018spider} define the SQL hardness based on the number of clauses and the number of components in a clause. Our method obtains larger gains on harder examples\footnote{The SQL query in ``Extra Hard'' level usually requires common knowledge and involves logical reasoning. The base model and our method are still far from resolving these.}, i.e, ``Medium'' and ``Hard'', in which 60\% of the SQL queries contain no less than three clauses. 

\textbf{Case study}. In order to verify the impact of clause-level alignments in the attention mechanism, we plot attention weights of original RATSQL and our RATSQL in Figure \ref{fig:matrix}. In our model, each clause has a higher attention weight with tokens in the corresponding aligned segment. Inversely, the base model doesn't have focus attention scores for some clauses, such as WHERE and GROUP, and it fails to generate the WHERE clause.

\section{Conclusion}
We propose clause-level parallel decoding and alignment loss to enhance grammar-based text-to-SQL parsing models. 
Experimental results show that our approach improves consistently both their efficiency and accuracy.


\bibliography{anthology,custom}
\bibliographystyle{acl_natbib}

\end{CJK}
\end{document}